\documentclass[letterpaper]{article} 
\usepackage[submission]{aaai24}  
\usepackage{times}  
\usepackage{helvet}  
\usepackage{courier}  
\usepackage{booktabs} 
\usepackage[hyphens]{url}  
\usepackage{graphicx} 
\usepackage{natbib}  
\usepackage{caption} 
\usepackage{lipsum}
\usepackage{xcolor}
\usepackage{amssymb}
\usepackage{algorithm}
\usepackage{algorithmic}

\usepackage{newfloat}
\usepackage{listings}
\DeclareCaptionStyle{ruled}{labelfont=normalfont,labelsep=colon,strut=off} 
\floatstyle{ruled}
\newfloat{listing}{tb}{lst}{}
\floatname{listing}{Listing}
\title{MCM: Multi-condition Motion Synthesis Framework for Multi-scenario}
\author{
	Zeyu Ling\textsuperscript{\rm 1}\equalcontrib 
	Bo Han\textsuperscript{\rm 1}\equalcontrib 
	Yongkang Wong\textsuperscript{\rm 2}
	Mohan Kangkanhalli\textsuperscript{\rm 2}
	Weidong Geng\textsuperscript{\rm 1}
}
\affiliations{
	\textsuperscript{\rm 1}Zhejiang University\\
	\textsuperscript{\rm 2}National University of Singapore\\
	
}

\usepackage{bibentry}

\begin{document}
	
	\maketitle
	\begin{figure*}[ht]
		\centering
		\includegraphics[width=0.95\textwidth]{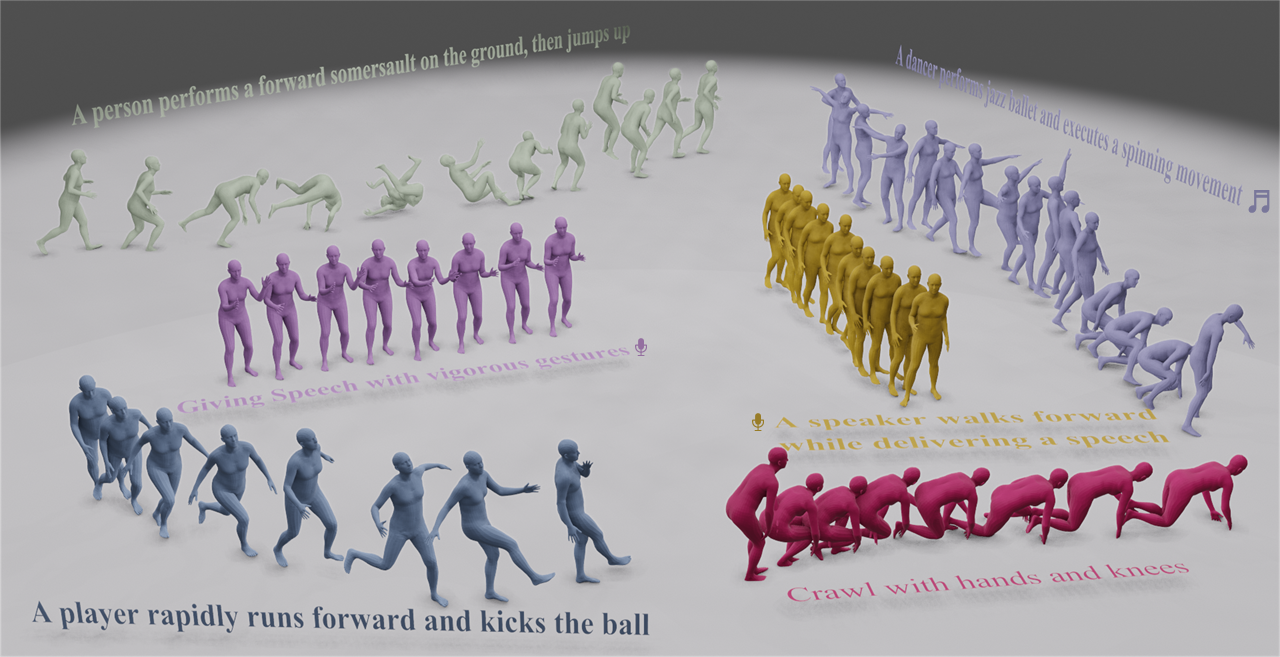}
		\caption{Our MCM method has generated human motion across various scenarios (e.g., text-to-motion or music-to-dance) based on different conditions (e,g, text, music, speech, etc.) By inputting challenging textual descriptions of actions such as kicking a ball, performing forward somersaults, crawling, and more, we have produced highly realistic sequences of movements. MCM is capable of generating motion sequences that not only align with rhythm but also match the dance descriptions(we use a musical note symbol to represent this scene). Additionally, MCM can generate co-speech motions based on speech audio and textual descriptions(a microphone note symbol).}
		\label{fig_main_vis}
	\end{figure*}
	
	\begin{abstract}
		
		The objective of the multi-condition human motion synthesis task is to incorporate diverse conditional inputs, encompassing various forms like text, music, speech, and more. This endows the task with the capability to adapt across multiple scenarios, ranging from text-to-motion and music-to-dance, among others. While existing research has primarily focused on single conditions, the multi-condition human motion generation remains underexplored.  
		In this paper, we address these challenges by introducing MCM, a novel paradigm for motion synthesis that spans multiple scenarios under diverse conditions. The MCM framework is able to integrate with any DDPM-like diffusion model to accommodate multi-conditional information input while preserving its generative capabilities.
		Specifically, MCM employs two-branch architecture consisting of a main branch and a control branch. The control branch shares the same structure as the main branch and is initialized with the parameters of the main branch, effectively maintaining the generation ability of the main branch and supporting multi-condition input. We also introduce a Transformer-based diffusion model MWNet (DDPM-like) as our main branch that can capture the spatial complexity and inter-joint correlations in motion sequences through a channel-dimension self-attention module.
		Quantitative comparisons demonstrate that our approach achieves SoTA results in both text-to-motion and competitive results in music-to-dance tasks, comparable to task-specific methods. 
		Furthermore, the qualitative evaluation shows that MCM not only streamlines the adaptation of methodologies originally designed for text-to-motion tasks to domains like music-to-dance and speech-to-gesture, eliminating the need for extensive network re-configurations but also enables effective multi-condition modal control, realizing "once trained is motion need". The code will be released at https://github.com/ZeyuLing/MCM.
		
	\end{abstract}
	
	\section{Introduction}
	
	Human motion generation finds extensive applications in fields such as film production, game development, and simulation. Traditional manual animation techniques are notably constrained in terms of efficiency. The emergence of neural network-based motion generation methods holds great promise and potential for enhancing the efficiency of motion generation.  However, achieving fine-fidelity human motion sequences remains a formidable challenge. 
	
	To address this issue, generative models including Variational Autoencoders (VAEs)~\cite{kingma2013auto}, Generative Adversarial Networks (GANs)~\cite{goodfellow2014generative}, Denoising Diffusion Probabilistic Models (DDPM)~\cite{ho2020denoising} have been adapted for human motion domain.
	
	Nonetheless, prevailing methods suffer from critical limitations. Firstly, they lack the ability to simultaneously handle multiple modal control conditions. For instance, certain approaches~\cite{guo2022generating,guo2022tm2t,zhang2022motiondiffuse,zhang2023t2m,chen2023executing} solely support textual conditions, while ~\cite{siyao2022bailando,tseng2023edge,li2021ai} only support music as conditions. Though some methods have demonstrated adaptability to multiple tasks, such as MDM~\cite{tevet2022human} for text-to-motion, motion editing and prediction, as well as MoFusion~\cite{dabral2023mofusion} for music-to-dance and text-to-motion tasks, they fail to handle multiple modalities of input concurrently, as each model only accepts a single modality of control conditions. Secondly, certain methods support multi-modal conditions but lack generalization capabilities for other scenarios. MultiContext~\cite{yoon2020speech} and CaMN~\cite{liu2022beat} accept different modalities of conditions like audio, text, and speaker ID, yet they are exclusively applicable to speech-to-gesture scenario, failing to exhibit versatility across diverse scenarios. These limitations constrict the applicability of current motion generation methods, limiting them to specific control conditions.
	
	To surmount these challenges, we propose a novel end-to-end framework \textbf{MCM} (\textbf{M}ulti-\textbf{C}ondition \textbf{M}otion synthesis framework for multi-scenario) based on the DDPM architecture, which is tailored for multi-scenario motion generation based on multiple conditions. Notably, our model adeptly accommodates diverse control conditions, including unprecedented combinations of conditions encountered outside the training set. For instance, by utilizing MCM, one can effectively describe dance motions with caption while providing background music, and the model can generate dance motions that synchronize with the music and align with the textual description. This obviates the need for constructing large datasets of caption-music-dance pairs, thereby alleviating the substantial burden of manual labor and economic resources required for dataset curation.
	
	MCM adopts a two-branch structure, comprising the main branch and the control branch. The main branch can leverage an arbitrary pre-trained DDPM network like MotionDiffuse~\cite{zhang2022motiondiffuse} and MDM~\cite{tevet2022human}, ensuring the quality and semantic coherence of the generated motions. On the other hand, the control branch initializes its parameters from the main branch and is responsible for providing fine-grained control capabilities, such as motion rhythm, style, etc.
	
	Additionally, prior works~\cite{zhang2022motiondiffuse,dabral2023mofusion,tevet2022human}, when employing attention modules, predominantly focused on modeling temporal and semantic-level information. However, when considering data modalities like motion sequences, it is imperative to recognize that the channel dimension holds valuable spatial information and inter-joint relationships within the human body, aspects that have often been underappreciated. Addressing this gap, we present MWNet, an innovative Transformer-Decoder architecture that integrates self-attention mechanisms tailored for the channel dimension. Our study substantiates the efficacy of this framework in the realm of motion generation.
	
	In summary, our core contributions are as follows:
	\begin{itemize}
		\item We introduce a unified framework MCM for multi-scenario motion generation based on multiple conditions. Remarkably, without necessitating structural reconfiguration of the network, MCM extends the capabilities of DDPM-based methods to accommodate additional conditional inputs.
		\item We propose a Transformer-Decoder architecture MWNet, enriched with a multi-wise attention mechanism, which adeptly leverages spatial information within motions.
		\item Exhaustive qualitative and quantitative assessment shows that our method outperforms existing methods in text-to-motion tasks and demonstrates competitive performance in music-to-dance tasks. Furthermore, our method exhibits favorable outcomes in novel scenarios involving multiple conditions.
	\end{itemize}
	
	\section{Related Work}
	
	Conditional human motion generation focuses on generating high-quality motion sequences that adhere to specific conditional constraints. The task encompasses various modalities of control conditions, leading to sub-tasks like text-to-motion, music-to-dance, motion prediction, motion interpolation, and speech-to-gesture.
	
	\subsection{Single-condition Human Motion Synthesis}
	
	Traditional VAE-based methods~\cite{guo2022generating,guo2022tm2t,petrovich2022temos,siyao2022bailando,ao2022rhythmic} typically involve two training stages: the encoder maps motion sequences to latent vector, while the decoder reconstructs the latent vector back into motion sequences. During the inference stage, after sampling the latent vector from the latent space, then reconstruct motion sequences with the guidance of conditions. DeepDance~\cite{sun2020deepdance}, MultiContext~\cite{yoon2020speech}, and DanceFormer~\cite{li2022danceformer} employ Generative Adversarial Networks to generate human motions. Due to the diversity and complexity of human motion, traditional VAE-based models cannot fully capture the distribution of human motion, while GAN-based methods often face the issue of mode collapse.
	Diffusion models have demonstrated remarkable efficacy across diverse tasks~\cite{rombach2022high,nichol2021glide,mei2023vidm}. Attributable to its stochastic nature, the diffusion model~\cite{ho2020denoising} is more suitable for modeling human actions with high diversity distribution features. MotionDiffuse~\cite{zhang2022motiondiffuse} and EDGE~\cite{tseng2023edge} separately used diffusion model in text-to-motion and music-to-dance. MDM~\cite{tevet2022human} uses the same network architecture to achieve multiple tasks, such as text-to-motion and motion edition. MAA~\cite{azadi2023make} pre-trained a diffusion model with a curated large-scale dataset of (text, static pseudo-pose) pairs extracted from image-text datasets, which significantly improves performance on captions outside of the distribution of motion capture datasets like ~\cite{guo2022generating,plappert2016kit,punnakkal2021babel}. MLD~\cite{chen2023executing} combines VAE (Variational AutoEncoder) with Diffusion Model and proposes the first latent space diffusion model in the field of motion generation. T2MGPT~\cite{zhang2023t2m} combines VQ-VQE and GPT~\cite{radford2018improving} for human motion generation from textural descriptions.
	
	
	\subsection{Multi-condition Human Motion Synthesis}
	
	Many efforts have been dedicated to the development of motion generation networks, aiming to accommodate various modalities of input. The GAN-based approach, MultiContext~\cite{yoon2020speech}, achieves the fusion of multiple modal conditions in the speech-to-gesture task. It utilizes speaker's voice, speech text, and speak ID as conditions to generate accompanying motions for speech. Building upon this foundation, CaMN~\cite{liu2022beat} introduces a more robust architecture that combines five distinct modalities as conditioning factors for generating accompanying speech actions: Speaker ID, speaker's emotion, speech text, speech sound, and speaker's facial expressions. However, both MultiContext and CaMN have not demonstrated the ability to generalize beyond the speech-to-gesture domain.
	
	MoFusion~\cite{dabral2023mofusion} is the first method that can handle diverse modal information and generalize across various scenarios. It's based on the diffusion model and capable of taking music or text as inputs, thus enabling tasks like text-to-motion or music-to-dance. However, this approach is incapable of simultaneously accepting both textual and auditory conditions as inputs. Therefore, fundamentally, it remains a single-condition generation model.
	
	While multi-condition, multi-scenario generation in the motion field is underexplored, ControlNet~\cite{zhang2023adding} has achieved highly effective image generation under multi-condition control. This novel approach allows for fine-grained control over generated images, utilizing conditions such as sketches and edge lines, in addition to the textual descriptions. This approach serves as a valuable source of inspiration for our proposed multi-condition framework.
	
	\section{Method}
	
	\subsection{Problem definition} 
	The objective of the human motion generation task is to generate a motion sequence $X \in \mathbb{R}^{T \times D}$ under a set of constraint conditions $C$. $X$ is an array of $x_{i}$, where $i \in \{1, 2, \ldots, T\}$, and $T$ denotes the number of frames. Each $x_{i} \in \mathbb{R}^{D}$ represents the D-dimensional pose state vector at the $i$-th frame. $c_{j} \in C$  could be textual description, speech voice, or background music.
	
	\begin{figure*}[t]
		\centering
		\includegraphics[width=0.95\textwidth]{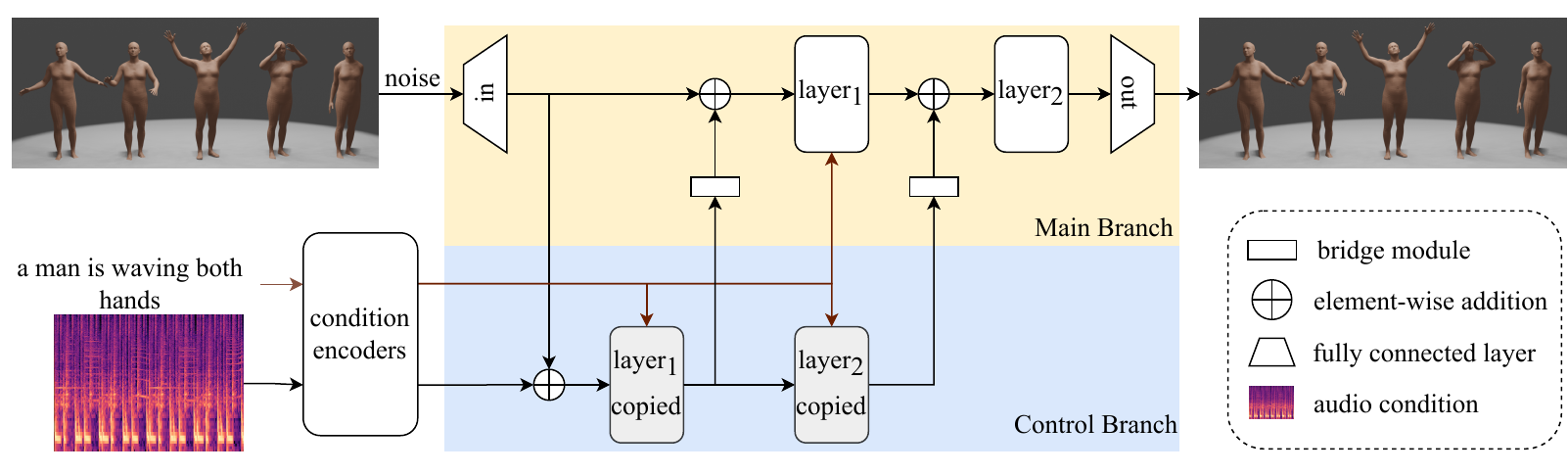} 
		\caption{MCM framework overview. MCM employs a dual-branch structure consisting of the main branch and the control branch. The layer wise outputs from the control branch are connected to the main branch via bridge modules, which are fully connected layers or 1d-convolutions with parameters initialized to zero. The output of each bridge module is directed added to the input feature vector of corresponding layers in the main branch. The condition encoders encompass several pre-trained feature extractors for different modal conditions. The fully connected layer ``in'' is responsible for mapping the motion vector to the hidden vector, while the ``out'' layer performs the opposite mapping.}
		\label{fig_overall_arch}
	\end{figure*}
	
	\subsection{MCM Framework} 
	An overview of the MCM Framework is described in Figure \ref{fig_overall_arch}. 
	We adopt a two-branch architecture consisting of a main branch and a control branch, and a two-stage training strategy to better incorporate multiple conditions. The main branch consists of arbitrary neural network layers, which can also be a pre-trained motion generation diffusion model, for instance, MotionDiffuse and MDM. The control branch shares the same structure as the main branch and is directly initialized with the parameters of the main branch.
	
	For each branch, we separately optimize them. The main branch is optimizing using text-motion paired data. If a pre-trained motion generation diffusion model is utilized, this stage can be omitted. The primary objective of this stage is to acquire text-to-motion correspondence.
	
	The second stage is denoted as the ``control'' stage. During this stage, all parameters, with the exception of those belonging to the control branch and bridge module, are set as fixed to ensure the preservation of the main branch's generation quality and semantic association capabilities. Control-motion-text paired data is employed in this phase, where the text data encompasses straightforward textual descriptions, such as ``a man dances Pop.'' or ``a man gives a speech.'' The control aspect can be represented by music, speech, or other control signals. Within the control branch, the output of each module is directly added to the corresponding original input of the main branch layers through the bridge module. This operation serves as the new input for the main branch layers, enabling the integration of control signals to guide human motion generation.

	\subsection{MWNet Architecture} 
	MCM establishes a framework for multi-conditional control to generate human motions. It can construct a main branch and a control branch based on any given DDPM-based model, allowing for the simultaneous processing of multi-conditional information. However, current DDPM-based motion generation models, such as MotionDiffuse and MDM, primarily focus on time-wise self-attention and cross-attention to model the time-level correlation and semantic-level correlation between motions and conditions. 
	
\begin{figure}[t]
		\centering
		\includegraphics[width=0.9\columnwidth]{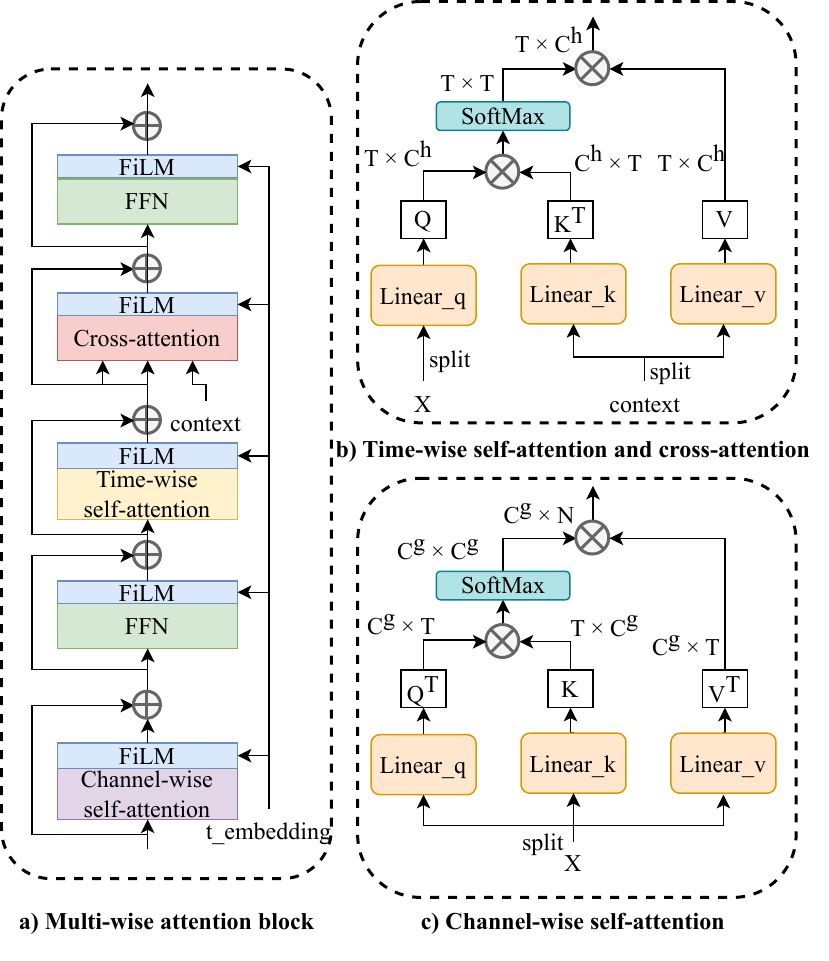} 
		\caption{Model architecture for a multi-wise attention block. It uses three types of attention modules alternatively. The symbols ``+'' and ``×'' separately represent feature addition and multiplication operation. $T$ symbolizes the length of the input sequence, while $C^g$ and $C^h$ signify the number of channels for the matrices $Q$, $K$, and $V$ after. The split operation means splitting the channels into $g$ groups or $h$ heads. Context represents text condition for cross-attention and is exactly equal to $X$ for time-wise self-attention.} 
		\label{fig_arch}
	\end{figure}

	However, motion sequences comprise positional and rotational information for each joint at every frame. In the context of a motion feature, the channel dimension encompasses spatial details and joint correlation of the motion, which is underexplored. Therefore, we opt for channel-wise self-attention~\cite{ding2022davit} and propose MWNet to model these crucial aspects of the information. MWNet consists of modified transformer decoder layers as shown in Figure \ref{fig_arch} (a), named as \textbf{M}ulti-\textbf{W}ise attention blocks.
	
	Similar to StableDiffusion~\cite{rombach2022high} and ~\cite{nichol2021glide}, we use FiLM~\cite{perez2018film} blocks to furnish timestamp information to MWNet after every attention or Feed Forward Network (FFN) layers. Output from a previous layer $x$ and timestamp embedding $\epsilon_t$ is given to a FiLM Block. The block processes the feature as follows:
	\begin{equation}
		FiLM(x, \epsilon_t)=x+LN(x\odot(W_1+I)\epsilon_t)+W_2\epsilon_t
	\end{equation}
	$LN$ denotes layer normalization layer~\cite{ba2016layer}. $W_1$, and $W_2$ are two projection matrices. $I$ represents a matrix with all elements being 1 and the shape is the same as $x$. $\odot$ denotes the element-wise multiplication.
	
	With projection weights $W^Q$, $W^K$, $W^V$, $X$ are projected to $Q=XW^Q$, $K=XW^K$, $V=XW^V$ and split into $N_h$ heads or $N_g$ groups. We denote $Q_i$, $K_i$, $V_i$ for each head or group. Time-wise self-attention can be denoted as follows:
	\begin{equation}
		SA_T(Q_i,K_i,V_i)=Softmax(\frac{Q_iK_i^T}{\sqrt{C_h}})V_i
	\end{equation}
	\begin{equation}
		SA_T(Q,K,V)=\{SA_T(Q_i,K_i,V_i)\}_{i=0}^{N_h}
	\end{equation}
	Whereas, channel-wise self-attention can be denoted as:
	\begin{equation}
		SA_C(Q_i,K_i,V_i)=(Softmax(\frac{Q_i^TK_i}{\sqrt{C_g}})V_i^T)^T
	\end{equation}
	\begin{equation}
		SA_C(Q,K,V)=\{SA_C(Q_i,K_i,V_i)\}_{i=0}^{N_g}
	\end{equation}
	
	$C_h$ and $C_g$ denotes the number of channels for each head or group.
	
	\section{Experiments}
	
	\subsection{Data preprocessing} 
	Following HumanML3D, we use a 263-dimension representation $x=concat(\dot{r}^a,\dot{r}^x,\dot{r}^z,r^y,j^p,j^v,j^r,c^f)$ to represents motions at every frame. $\dot{r}^a \in R$ is root angular velocity along Y-axis; $\dot{r}^x,\dot{r}^z \in R$ are root linear velocities on XZ-plane; $r^y$ is root height; $j^p \, j^v \in R^{3j}$ and $j^r \in R^{6j}$ are the local joints positions, velocities, and rotations in root space, with $j$ denoting the number of joints; $c^f \in R^4$ is binary features obtained by thresholding the heel and toe joint velocities to emphasize the foot ground contacts. To train on various datasets, we process all datasets to the same format with a 22-joint skeleton (the first 22 joints of the SMPL skeletal structure) and 20 FPS. 
	
	\subsection{Implement Details} 
	We conduct training of MCMs utilizing distinct DDPM-like main branch architecture, including MotionDiffuse~\cite{zhang2022motiondiffuse}, MDM~\cite{tevet2022human}, and our MWNet. The conditioning inputs from diverse modalities are pre-processed through the employment of pre-trained condition encoders. For encoding textual prompts, we employ the CLIP-base pretrained model~\cite{radford2021learning}. In handling audio conditions, we leverage Jukebox~\cite{dhariwal2021diffusion} for music processing and HuBERT-base~\cite{hsu2021hubert} for vocal processing. Subsequently, the resultant feature vectors are projected onto a common dimension and concatenated. This portion will be elaborated on in the supplementary materials. Regarding the diffusion model, we set the number of diffusion steps at 1000, while the variances $\beta_t$ follow a linear progression from 0.0001 to 0.02. We employ the Adam optimizer for training the model, employing a learning rate of 0.0002 throughout both training phases.
	
	\subsection{Text-to-Motion Generation} 
	We train and evaluate our main branch model MWNet on HumanML3D~\cite{guo2022generating} dataset. It consists of about 28k motions, each with 3 or 4 captions. The metrics are similar to prior works~\cite{zhang2022motiondiffuse}: Frechet Inception Distance (FID), Top-k R-Precision, MultiModal Distance, Diversity, and MultiModality. 
	
	\subsubsection{FID}	
	With a pre-trained encoder to extract feature vectors from generated motion and real motion respectively, FID evaluates the dissimilarity between two distributions by calculating the difference between feature vector statistical measures (mean and covariance).
	
	\subsubsection{Diversity}
	The diversity metric calculates the average pairwise Euclidean distance among random pairs in the dataset, irrespective of input prompts.
	
	\subsubsection{Top-k R-Precision}
	The R-Precision score assesses the classification accuracy of generated motions using a pre-trained classifier~\cite{guo2022generating}. It quantifies how often the top-k closest motions in Euclidean distance to their corresponding captions are achieved within a 32-sample batch. 
	
	\subsubsection{MultiModal Distance}
	The computation of the MultiModal Distance metric involves the use of a pair of pre-trained feature extractors, trained via contrastive learning, to extract features from generated motions and target captions. The distance between these features is then calculated. A smaller MultiModal Distance typically indicates a strong match between the two modalities.
	
	\subsubsection{MultiModality}
	It gauges diversity by sampling the method N times, which calculates the average pairwise Euclidean distance of generated motions from the same text input, where a greater distance indicates higher variability.

	\begin{table*}[t]
		\centering
		\small
		\begin{tabular}{lccccccc}
			\toprule
			Methods & \multicolumn{3}{c}{R Precision $\uparrow$} & FID $\downarrow$ & MultiModal Dist $\downarrow$ & Diversity $\rightarrow$ & MultiModality $\uparrow$ \\
			& Top 1 & Top 2 & Top 3 & & & & \\
			\midrule
			Real motions & 0.511 & 0.703 & 0.797 & 0.002 & 2.974 & 9.503 & - \\
			\midrule
			T2M et al. & 0.457 & 0.639 & 0.740 & 1.067 & 3.340 & 9.188 & 2.090 \\
			T2MGPT($\tau$=0) & 0.417 & 0.589 & 0.685 & 0.140 & 3.730 & 9.844 & \textcolor{red}{3.285} \\
			T2MGPT($\tau$=0.5) & 0.491 & 0.680 & 0.775 & \textcolor{blue}{0.116} & 3.118 & 9.761 & 1.856 \\
			T2MGPT($\tau \in \mathcal{U}[0, 1]$) & \textcolor{blue}{0.492} & 0.679 & 0.775 & 0.141 & 3.121 & 9.722 & 1.831 \\
			\midrule
			MLD &- &- &0.772 &0.473 &3.196 &9.724 &2.413 \\
			MotionDiffuse & 0.491 & \textcolor{blue}{0.681} & \textcolor{red}{0.782} & 0.630 & \textcolor{blue}{3.113} & 9.410 & 1.553 \\
			MDM & - & - & 0.611 & 0.544 & 5.566 & \textcolor{blue}{9.559} & \textcolor{blue}{2.799} \\
			MoFusion & - & - & 0.492 & - & - & 8.820 & 2.521 \\
			\textbf{MWNet(ours)} & \textcolor{red}{0.494} & \textcolor{red}{0.682} & \textcolor{blue}{0.777} & \textcolor{red}{0.075} & \textcolor{red}{3.086} & \textcolor{red}{9.484} & 0.968 \\
			\bottomrule
		\end{tabular}
		\caption{Quantitative results on the HumanML3D test set. All methods use the real motion length from the ground truth. $\rightarrow$ means results are better if the metric is closer to the real distribution(metrics of real motions). The method highlighted in bold font is based on the Diffusion Model. Methods below MotionDiffuse(including MotionDiffuse) are based on DDPM, while the others are not. We use the red font to highlight the metric of the first position and blue for the second.}
		\label{table_eval_text2motion}
	\end{table*}

	Table \ref{table_eval_text2motion} presents the quantitative metrics of our method on the HumanML3D dataset. In terms of FID, MultiModal Dist, Diversity, R-precision top 1, and top 2 metrics, MWNet has achieved state-of-the-art results. MWNet ranks second only to MotionDiffuse in terms of R-precision Top 3. We believe that such remarkable performance is attributed to the Multi-wise attention mechanism we have employed. We will delve deeper into this in the supplementary materials.
	
	\subsection{Music-to-Dance Generation} 
	After the main branch model training stage on HumanML3D Dataset, we proceed with the control branch training on the AIST++ dataset~\cite{li2021ai}. This dataset encompasses 1408 distinct dance motion sequences, spanning durations from 7.4 to 48.0 seconds. It encompasses ten distinct dance motion genres, each featuring multiple dance choreographies within its genre. This intricate arrangement fosters a substantial diversity, encompassing a wide spectrum of dance motions. Based on the dance motion descriptions provided by AIST++, we generated pseudo-captions to serve as textual inputs for MCM. For example, ``A male dancer performs Pop in Cypher to music,'' accompanied by comprehensive details encompassing the dancer's gender (male, female), dance genre (Pop, Break, etc.), and dance context (group dance, showcase, Cypher, etc.).
	
	We conduct the quantitative evaluation for music-conditioned motion generation using evaluation metrics following~\cite{dabral2023mofusion}. (1) FID: utilizing kinetic features~\cite{onuma2008fmdistance} implemented within fairmotion~\cite{gopinath2020fairmotion}. The kinetic feature extractor transforms body joint positions $X \in R^{T \times J \times 3}$ into kinetic features $z_k \in R^{3J}$. Here, $T$ represents the number of frames, and $J$ signifies the number of joints. (2) Diverisy: it computes the average pairwise Euclidean distance of the kinetic features of the motions generated from music in the test set. (3) Beat Alignment Score (BAS): a metric that quantifies the congruence between kinematic beats and musical beats. Kinematic beats correspond to the local minima of kinetic velocity within a motion sequence, signifying points where motion momentarily halts. Additionally, we extract music beats from the audio signal utilizing the Librosa~\cite{mcfee2015librosa} toolbox. The BAS is computed as the average distance between each music beat and its nearest dance beat:
	\begin{equation}
		BAS=\frac{1}{|B^m|}\sum_{t^m \in B^m}^{}exp\{-\frac{min_{t^d \in B^d}||t^d-t^m||^2}{2\sigma^2}\}
	\end{equation}
	
	$B^d$ represents the beat timings within dance motions, and $B^m$ corresponds to the beat timings in the music. The parameter $\sigma$ is a normalized value, in line with Bailando~\cite{siyao2022bailando}, which is set to 3 in our experiments.
	\begin{figure}[t]
		\centering
		\includegraphics[width=0.9\columnwidth]{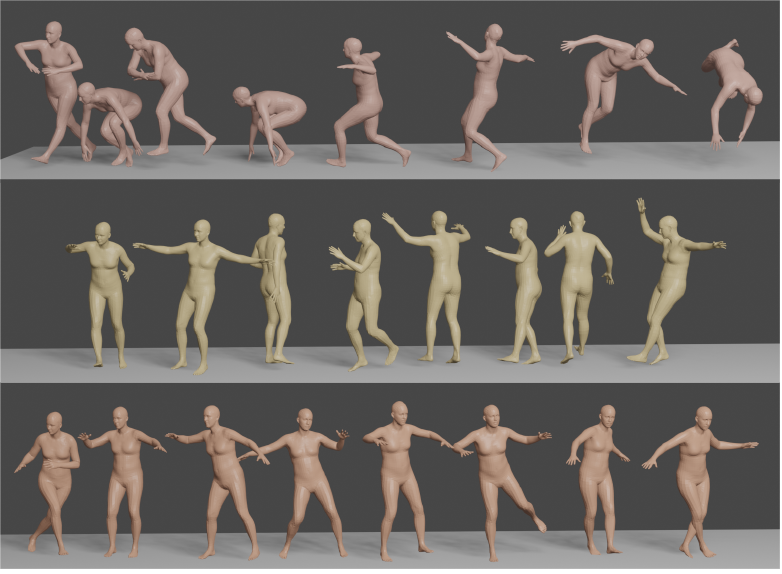}
		\caption{Dance genre control with different text prompts. From top to bottom, using the same piece of music, we input text descriptions ``A dancer performs Break'', ``Waack'', and ``Lock'' in addition to music.}
		\label{fig_dance_genre}
	\end{figure}
	As the same as in HumanML3D, we slice the AIST++ dataset into segments of up to 10 seconds, with a frame rate of 20 FPS, and process them into the previously mentioned 263-dimensional vector representations. All the methods compared are trained on the AIST++ training set and evaluated on the validation and test sets. 
	
	Table \ref{table_eval_music2dance} showcases the performance results of our method using the AIST++ dataset. To the best of our knowledge EDGE~\cite{tseng2023edge} is the only open-source task-specific music-to-dance method based on the diffusion model and achieves state-of-the-art performance on the AIST++ dataset, with the highest Beat Align score and second highest diversity. Our results outperform EDGE in all metrics. The diversity of dance movements generated by our three models surpasses that of EDGE. MWNet+MCM and MDM+MCM achieved a bit lower FID scores than EDGE. MWNet+MCM achieved Beat Align Scores similar to EDGE. It's worth noting that the MCM-based method was trained on the AIST++ training set for no more than 1000 epochs, while EDGE was trained for about 8000 epochs. We believe this is attributed to our two-stage training strategy. The MCM-based method acquired the ability to generate high-quality motions during the first-stage training for the text-to-motion task. Therefore, in the second stage, it required fewer epochs to converge rapidly. Based on these findings, we believe our methods are capable of generating dance movements comparable to task-specific dance generation methods. 
	\begin{table}[htb]
		\centering
		\begin{tabular}{lccc}
			\toprule
			Methods & FID & Div & BAS \\
			\midrule
			Real Motions & - & 9.636 & 0.314 \\
			\midrule
			EDGE & 39.584 & 5.754 & 0.274 \\
			MotionDiffuse + MCM  & 51.929 & 10.453 & 0.246 \\
			MDM + MCM & 39.434 & 7.157 & 0.265 \\
			\textbf{MWNet + MCM} & 38.251 & 8.296 & 0.275 \\
			\bottomrule
		\end{tabular}
		\caption{Results on AIST++ validation and test set. }
		\label{table_eval_music2dance}
	\end{table}

	\begin{figure}[t]
		\centering
		\includegraphics[width=0.9\columnwidth]{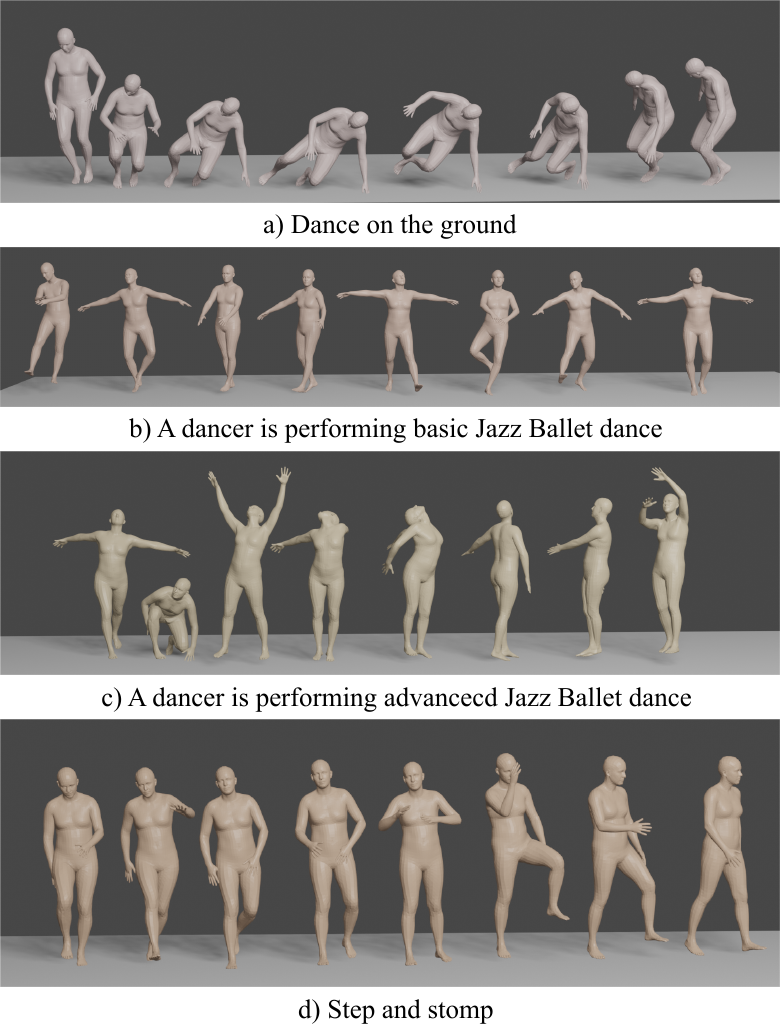}
		\caption{Dance details control with different text prompts}
		\label{fig_dance_control}
	\end{figure}
	
	\subsection{Multi-condition Generation} 
	We conduct extensive multi-condition controlled human motion generation experiments. As shown in Figure \ref{fig_dance_genre}, we use the same piece of music and different text prompts to control the genre of generated dance motions. we input text descriptions “A dancer performs Break”, “Waack”, and “Lock” sequentially to guide MCM in generating dance movements of different genres. The distinctive feature of Break is often its incorporation of ground movements, Lock frequently involves body locks and control, while Waack emphasizes arm movements. 
	
	In Figure \ref{fig_dance_control}, we demonstrate the fine-grained control of dance movements by MCM. Under the same piece of music, we use textual description to control various aspects of the dance movements, including specific dance movements and levels of difficulty. With the textual description, we control the specific actions and difficulty of the dance. In (a), we use a piece of Break style music and request a dance involving floor movements. In (b) and (c), based on the same Jazz Ballet style music, we generate relatively simple basic jazz ballet movements and more challenging advanced jazz ballet movements, including jumps and fast spins. In (d), given Waack-style music, we ask for dance movements involving kicking and stomping.
	
	We also conduct the second training stage (control branch) on BEAT dataset~\cite{liu2022beat} for the speech-to-gesture task. We fit the motion sequences provided by the BEAT dataset using the SMPL-X~\cite{pavlakos2019expressive} model, selecting the necessary 22 key points and transforming them into the 263-dimensional vector representation. Simultaneously, we slice the motions in the dataset into segments of up to 10 seconds at 20 FPS.  We use text prompts constructed from the speech of a speaker, including the speaker's voice and spoken content (e.g., A male speaker is saying: ``I am shocked by what you have done.''), as conditions for generating motions. By adjusting the text prompts, we can change the specific movements and amplitude when the person is speaking, such as waving, nodding, and more. 
	
\begin{figure}[t]
		\centering
		\includegraphics[width=0.9\columnwidth]{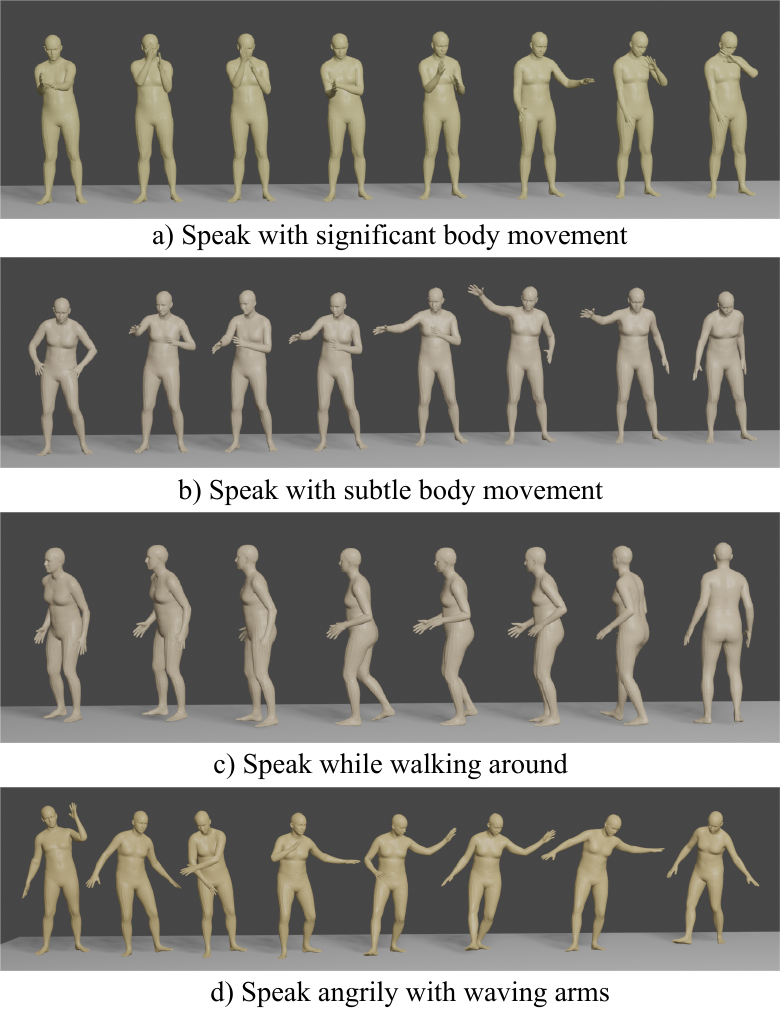}
		\caption{Speech action control with different text prompts}
		\label{fig_s2g}
	\end{figure}

	As shown in Figure \ref{fig_s2g}, by inputting different descriptive texts along with the same audio of a person's voice, we obtain varying accompanying actions. In (a) and (b), we task the MCM to generate subtle and significant accompanying actions respectively, and it's evident that the person's motions in (b) are noticeably more pronounced. In (c), we provide the description ``speak while walking around on the stage.'' In (d), we employ the description ``A man is speaking angrily with arms waving'' to generate a sequence of actions conveying an angry speaking gesture. Under the same segment of human voice audio, by modifying the input text, we can generate various distinct accompanying speech actions. Additionally, we can exert fine-grained control over the intensity, emotion, and movement aspects of the actions. 
		
	\section{Conclusion}
	We propose MCM, a novel paradigm for the multi-conditioned motion generation method that spans multiple scenarios. With MCM, DDPM-like methods designed for text-to-motion can simultaneously accommodate multiple modal conditions without requiring any structural adjustments. Additionally, we introduce a Transformer-based architecture MWNet that incorporates channel-wise self-attention, enhancing the modeling of spatial information and inter-joint correlations. We quantitatively evaluate our approach across tasks based on various modal conditions. In text-to-motion tasks reliant on text inputs, our method demonstrates superiority over other existing approaches. We further conducted qualitative assessments on tasks involving simultaneous multi-modal inputs, encompassing text-controlled music-dance generation and speech gesture synthesis. These tests demonstrated MCM's capability to generate actions under various control conditions.

	\bibliography{aaai24}
	
\end{document}


\maketitle
\begin{table*}
	\centering
	\begin{tabular}{lccccccc}
		\toprule
		Methods & \multicolumn{3}{c}{R Precision↑} & FID↓ & MultiModal Dist↓ & Diversity→ & MultiModality↑ \\
		& Top 1 & Top 2 & Top 3 & & & & \\
		\midrule
		channel-post+epsilon & 0.280 & 0.443 & 0.559 & 4.051 & 4.568 & 7.895 & 3.787 \\
		channel-first+epsilon & 0.378 & 0.565 & 0.680 & 1.256 & 3.792 & 8.515 & 2.244 \\
		channel-post & 0.455 & 0.642 & 0.744 & 0.751 & 3.399 & 8.933 & 1.707 \\
		channel-first & 0.455 & 0.640 & 0.732 & 0.377 & 3.349 & 9.312 & 1.481 \\
		\bottomrule
		
	\end{tabular}
	\caption{Comparison of performance under different MWNet layouts and training loss combinations. Methods with "+epsilon" in their names calculate the loss between the predicted noise and the true noise, while those without it predict $x_0$ and compute the loss against the real motion sequence. }
	\label{table_text2motion_diff}
\end{table*}

\begin{figure}[t]
	\centering
	\includegraphics[width=0.9\columnwidth]{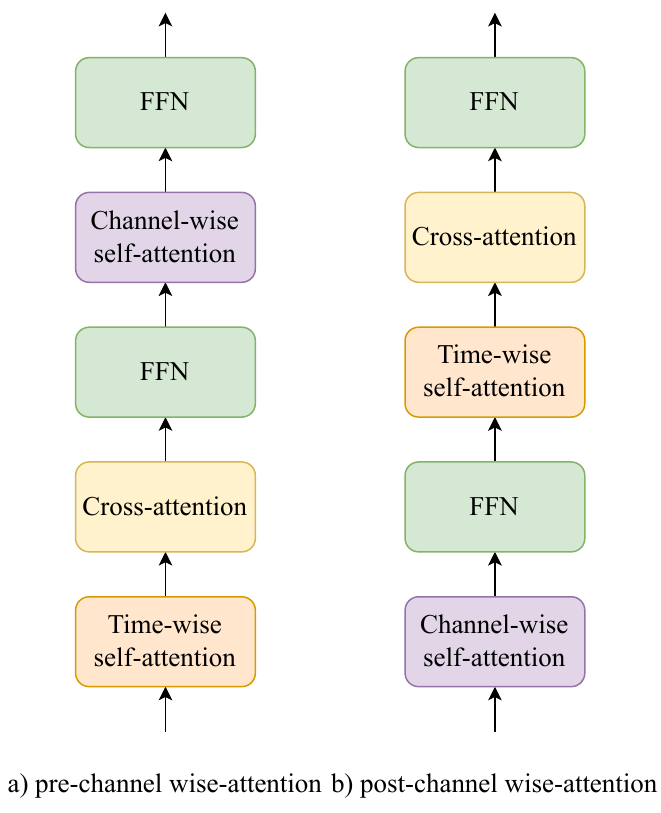} 
	\caption{MWNet layouts under different orders of attention. We omitted the FiLM~\cite{perez2018film} module after each module.}
	\label{fig_layout}
\end{figure}

\begin{figure}[t]
	\centering
	\includegraphics[width=0.9\columnwidth]{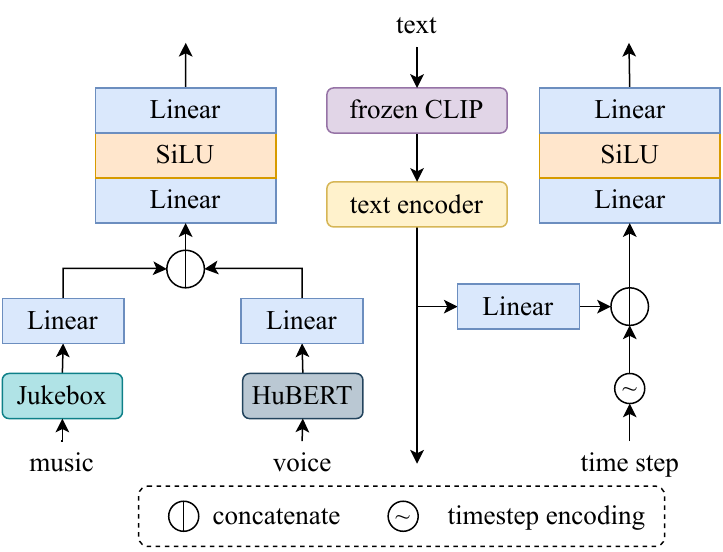} 
	\caption{Condition encoders in MCM.}
	\label{fig_condition_encoder}
\end{figure}

In the attachment we submitted, we have included the complete code for functionalities such as model, training, evaluation, and prediction. We have also provided the pre-trained checkpoint, which is located in the ``MCM'' directory within the attachment folder. In the ``MCM/data'' directory, we have included instructions and scripts for data processing. In ``demo gifs'' folder, we present some visualization demos for various scenarios.
\section{Data Preprocess}
This section introduces how we processed the raw dataset into the format we required.

In our approach, we processed the motion sequences from all the datasets used into a 263-dimensional vector representation as proposed in HumanML3D~\cite{guo2022generating} (for convenience, we will refer to this representation as "motion vector" in the following text). Additionally, we downsampled the frame rate of the motion to 20fps. 

For the HumanML3D dataset, we followed its official steps entirely. For the AIST++~\cite{li2021ai} dataset, we initially converted the motion data provided in SMPL format into motion vectors. Subsequently, both the music and motion vectors were downsampled to 20fps (16000 Hz). Slicing was then performed with a 1-second stride and a maximum segment length of 10 seconds, resulting in approximately 5000 dance-music training pairs and 48 test pairs. The BEAT~\cite{tseng2023edge} dataset supplied motion data in bvh format. We fitted the data using the SMPLX~\cite{pavlakos2019expressive} model (this process is not entirely reliable, and upon inspection, some samples exhibited distortions), generating motion data in SMPL format. Subsequently, we followed the same steps as with the AIST++ dataset.

In addition to sound and motion sequences, both AIST++ and BEAT provide some extra motion-related information. We utilize this information to generate pseudo-captions for training in the second stage (control stage). For instance, AIST++ offers additional details for each dance segment, such as dancer gender, dance genre, and dance occasion. Using this information, we generate sentences for descriptions, like: ``A male dancer followed the music and performed a Break dance in a Cypher.'' In the case of the BEAT dataset, it provides the gender of each speaker and the words in each speech segment. We concatenate these words to form a complete sentence, such as: ``A male speaker delivered the following speech,'' followed by the specific speech content.

\section{Detailed Implementation}

\textbf{Architecture and training loss} In Figure \ref{fig_layout}, we illustrated the variations in the layout of the multi-wise attention module when different orders of attention modules were employed. The sequence of different attention modules might influence the final model performance. Here, ``channel-first'' denotes the layout with the channel-wise self-attention placed in the earlier positions, while "channel-last" refers to its placement in the later positions. Experimental results from DaViT~\cite{ding2022davit} indicated that in the image domain, the impact of different layouts was marginal. However, within the domain of motion generation, our results highlighted this effect as significant.

Regarding training loss, similar to most DDPM~\cite{ho2020denoising} models, we utilized the simplest MSE loss. Nonetheless, we conducted comparisons on the entities over which the loss was computed. MotionDiffuse~\cite{zhang2022motiondiffuse} calculates the loss between the predicted noise and true noise at each timestep, whereas MDM~\cite{tevet2022human} directly predicts $x_0$ at each timestep and computes loss against the actual motion sequence. We experimentally verified how these two loss calculation methods impact motion generation tasks.

In Table \ref{table_text2motion_diff}, we compared the performance of MWNet in text-to-motion under different combinations of layouts and loss computation methods. To ensure fair comparison while avoiding excessively prolonged training times, each model was trained for only 500 epochs. 

Based on our experimental results, we can draw the following conclusions:
\begin{itemize}
	\item Directly predicting $x_0$ significantly enhances the quality of generated actions compared to predicting noise. Both methods for direct $x_0$ prediction exhibit superior performance in terms of action quality and semantic relevance compared to noise prediction. While in the field of image generation, the Diffusion Model often employs noise prediction, our experimental findings suggest that for action generation, the practice of directly predicting $x_0$ is more suitable.
	\item Placing the channel-wise self-attention at the front of the module effectively enhances the quality of action generation. Regardless of predicting $x_0$ or noise, layouts that position the channel-wise self-attention towards the front consistently exhibit improved action quality. Particularly when predicting noise, the channel-post method outperforms the channel-first method across all metrics.
\end{itemize}

\textbf{Condition encoders} We employed several types of condition encoders for the conditioning of different modalities.

As shown in \ref{fig_condition_encoder}, we utilized Hubert~\cite{hsu2021hubert} and Jukebox~\cite{dhariwal2020jukebox} to extract features from vocals and music, respectively. In our current application scenario, these two conditions do not coexist. However, in anticipation of potential simultaneous presence of both conditions in future work, we adopted the design as depicted in the diagram. Features from both modalities are concatenated together after being mapped to the same dimension, serving as joint audio features. If one of the modalities is absent, it will be substituted with an embedding of the same dimension.

For the textual condition, we employed a frozen CLIP~\cite{radford2021learning} module alongside an adaptable text encoder. In our implementation, the text encoder is a 4-layer Transformer. The resulting feature vectors are utilized in two ways: 1) for performing cross-attention operations with action features, and 2) for extracting the feature corresponding to the EOS token as a global semantic feature, which is then merged with timestep information.

\section{Qualitative Results}

In this section, we present qualitative results that were not feasible to display within the main body of the text. 

\begin{figure}[H]
	\centering
	\includegraphics[width=0.9\columnwidth]{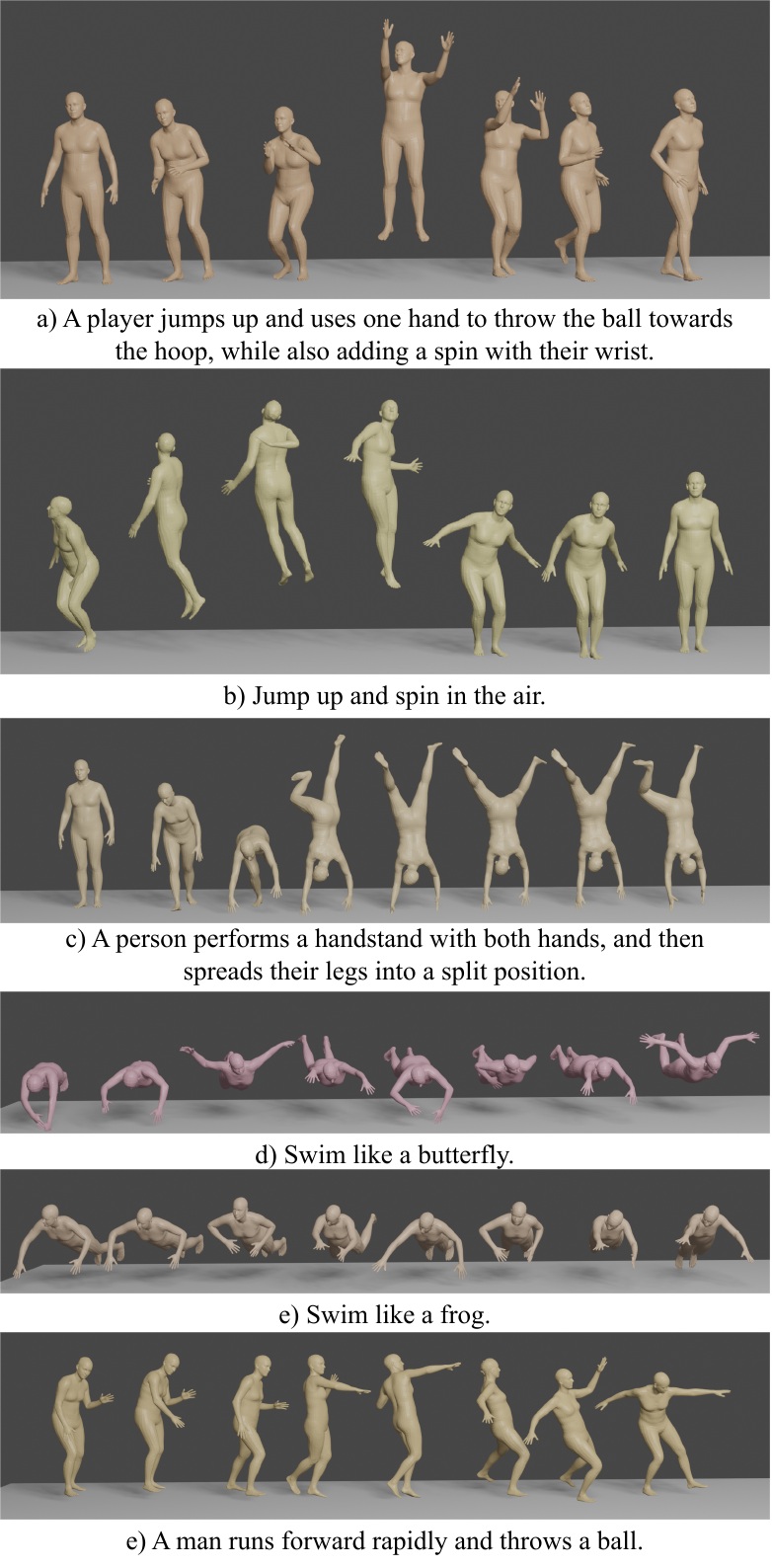} 
	\caption{We demonstrate the powerful semantic alignment capability of MWNet, which produces realistic effects even in challenging and complex motion scenarios.}
	\label{fig_text2motion}
\end{figure}

\textbf{text-to-motion} In figure \ref{fig_text2motion} We present qualitative results of our method in the text-to-motion task. For challenging and complex movements, our approach still demonstrates a strong semantic alignment and authenticity with the descriptive text.

\begin{figure}[H]
	\centering
	\includegraphics[width=0.9\columnwidth]{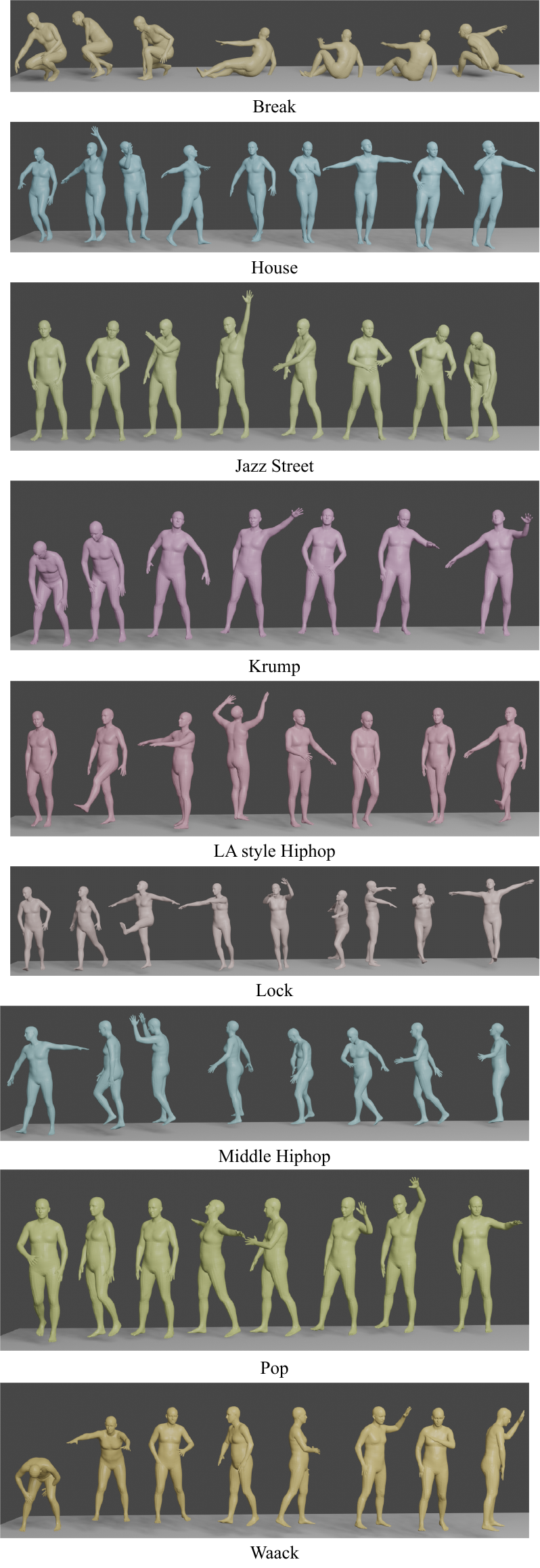} 
	\caption{We use the same textual description, ``a dancer performs advanced dance'', along with various styles of music to showcase MCM's capability to perceive different musical styles.}
	\label{fig_only_music}
\end{figure}

\textbf{music-to-dance} In the main body of the text, we primarily demonstrate MCM's capability to generate music under the joint control of text and music. In Figure \ref{fig_only_music}, we utilize the same textual description ``A dancer performs advanced dance'' and input various styles of music to verify MCM's ability to capture musical style information.

\bibliography{aaai24}